\documentclass[final]{cvpr}

\usepackage{times}
\usepackage{epsfig}
\usepackage{graphicx}
\usepackage{amsmath}
\usepackage{amssymb}
\usepackage{amsthm}
\usepackage{bm}
\usepackage{multirow}
\usepackage{multicol}
\usepackage{booktabs}
\newcommand{\tablestyle}[2]{\setlength{\tabcolsep}{#1}\renewcommand{\arraystretch}{#2}\centering\footnotesize}

\DeclareMathOperator*{\argmin}{arg\,min}
\usepackage[pagebackref=true,breaklinks=true,colorlinks,bookmarks=false]{hyperref}

\def\cvprPaperID{****} % *** Enter the CVPR Paper ID here
\def\confYear{CVPR 2021}

\begin{document}

%%%%%%%%% TITLE
\title{Unrestricted Adversarial Attacks on ImageNet Competition}

\author{Yuefeng Chen$^{1}$, Xiaofeng Mao$^{1}$, Yuan He$^{1}$, Hui Xue$^{1}$, Chao Li$^{1}$, Yinpeng Dong$^{2,3}$, \\ Qi-An Fu$^{2}$, Xiao Yang$^{2}$, Wenzhao Xiang$^{4}$, Tianyu Pang$^{2}$, Hang Su$^{2}$, Jun Zhu$^{2,3}$,
\\Fangcheng Liu$^{5}$, Chao Zhang$^{5}$, Hongyang Zhang$^{6}$, Yichi Zhang$^{2}$, \\ Shilong Liu$^{2}$, Chang Liu$^{2}$, Wenzhao Xiang$^{2}$, Yajie Wang$^{7}$,  Huipeng Zhou$^{7}$, \\ Haoran Lyu$^{7}$, Yidan Xu$^{7}$, Zixuan Xu$^{7}$, Taoyu Zhu$^{8}$, Wenjun Li$^{8}$, Xianfeng Gao$^{7}$, \\Guoqiu Wang$^{9}$,  Huanqian Yan$^{9}$,  Ying Guo$^{9}$, Chaoning Zhang$^{10}$, Zheng Fang$^{11}$, Yang Wang$^{11}$, \\ Bingyang Fu$^{11}$, Yunfei Zheng$^{11}$, Yekui Wang$^{11}$, Haorong Luo$^{11}$ and Zhen Yang$^{11}$\\
$^{1}$ Alibaba Group \; $^{2}$ Tsinghua University \; $^{3}$ RealAI \; $^{4}$ Shanghai Jiao Tong University \\ $^{5}$ Peking  University \; $^{6}$ University of Waterloo \; $^{7}$ Beijing Institute of Technology \\ $^{8}$ Guangzhou University \; $^{9}$ Beihang University \; $^{10}$ KAIST \; $^{11}$ Army Engineering University of PLA \\
\small{\{yuefeng.chenyf, mxf164419, heyuan.hy, hui.xueh, lizhao.lz\}@alibaba-inc.com} \\
\small{\{dyp17, qaf19, yangxiao19, pty17\}@mails.tsinghua.edu.cn}, \small{\{suhangss, dcszj\}@tsinghua.edu.cn}
} 

\maketitle

%%%%%%%%% ABSTRACT
\begin{abstract}
   Many works have investigated the adversarial attacks or defenses under the settings where a bounded and imperceptible perturbation can be added to the input. However in the real-world, the attacker does not need to comply with this restriction. In fact, more threats to the deep model come from unrestricted adversarial examples, that is, the attacker makes large and visible modifications on the image, which causes the model classifying mistakenly, but does not affect the normal observation in human perspective. Unrestricted adversarial attack is a popular and practical direction but has not been studied thoroughly. We organize this competition with the purpose of exploring more effective unrestricted adversarial attack algorithm, so as to accelerate the academical research on the model robustness under stronger unbounded attacks. The competition is held on the TianChi platform (\url{https://tianchi.aliyun.com/competition/entrance/531853/introduction}) as one of the series of AI Security Challengers Program.
  
\end{abstract}

%%%%%%%%% BODY TEXT
\section{Introduction}

%Recent years, with the vigorous technological development, AI has gradually become the driver of many practical applications. Meanwhile, the AI security also has become more important. Ensuring the AI security is challenging. As the AI defender in the future, Alibaba Security unions Tsinghua University and University of Illinois at Urbana-Champaign (UIUC) to convene "Challengers" for the safety of AI models based on different data modalities, such as text, image, video, audio and so on. They hope work together to build a safer AI.

Deep neural network (DNNs) has achieved the most advanced performance in various visual recognition problems. Despite its great success, the security of deep models have also caused many concerns in the industry. For example, DNNs are vulnerable to small and imperceptible perturbations on the inputs (these inputs are also called adversarial examples~\cite{szegedy2013intriguing,goodfellow2015explaining}). In addition to the small and imperceptible perturbations, in the actual scene, more threats to the deep model come from the unrestricted adversarial examples~\cite{Shamsabadi2020colorfool,Shamsabadi2020edgefool,duan2020adversarail,Bhattad2020unrestricted,guo2020watch}, that is, the attacker makes large and visible modifications on the image, which causes the model classifying mistakenly, but does not affect the normal observation in human perspective. Unrestricted adversarial attack is a popular direction in the field of adversarial attack in recent years. 
We hope that this competition can lead competitors not only understand and explore the scene of unrestricted adversarial attack on ImageNet, but also further refine and summarize some innovative and effective schemes of unrestricted attack, so as to promote the development of the field of adversarial attack academically.

In ImageNet classification task~\cite{imagenet}, in recent years, a variety of derived datasets (such as ImageNet-C~\cite{Hendrycks2019benchmarking}, ImageNet-A~\cite{Hendrycks2019natural}, ImageNet-Sketch~\cite{wang2019learning}, ImageNet-R~\cite{Hendrycks2020the}, etc.) have emerged to evaluate the robustness of various models in different scenarios. These datasets do not limit the similarity between the modified image and the original image, but require the attack image to be closer to the real world noise (image noise often occurs in practice like corruption, or out of distribution). The purpose of this competition is to explore a more realistic way to generate unrestricted adversarial examples. At the same time, through a variety of unrestricted attack testing, it helps us to understand the vulnerability of the current deep models and build a more robust image classification service.

The competition uses Imagenet dataset~\cite{imagenet}, we selected 15000 images, the whole competition is divided into three stages: first, we open 5000 images for players to debug the algorithm, the samples submitted by players will be scored by a series of pre-defined metrics; after the first stage, we open another 5000 images, and go into the second stage, in this stage the scoring metrics is consistent with stage 1, but we replaced the deployed image recognition model. The second stage lasted only 7 days; After the end of the second stage, we will select the teams that enter the final stage through two principles: 1. According to the objective score of the second stage, the top 10 teams will enter the final stage; 2. According to the objective score of the second stage, the top 11-30 teams will be recommended anonymously by the experts of the organizing committee, and the top 10 teams will enter the final stage. In the final stage, players need to submit the model and running environment for generating adversarial examples. We will test on the remaining 5000 private images and use human to score. We will manually mark the score of the submitted examples by the marking staff. The score marking is considered from two dimensions: first, the changed image needs to have the same semantics as the original image and guarantee the singleness of the semantics; second, the image needs to have higher visual quality, and human can clearly identify the object. The final result of the competition is decided by the final stage.

\section {Unrestricted Adversarial Attacks on ImageNet}

In this field, various great works are proposed to perform unrestricted adversarial attacks from different perspectives. Shamsabadi \textit{et al.}~\cite{Shamsabadi2020edgefool} propose an adversarial image enhancement filter termed EdgeFool which crafts adversarial examples with enhanced details. Bhattad \textit{et al.}~\cite{Bhattad2020unrestricted} present two interesting semantic attacks which manipulate texture and color fields to generate adversarial examples. Guo \textit{et al.}~\cite{guo2020watch} investigate the potential hazards of motion blur for DNNs, and propose kernel-prediction-based attack which makes the motion-blurred adversarial examples visually more natural. Shamsabadi \textit{et al.}~\cite{Shamsabadi2020colorfool} exploit image semantics
to selectively modify colors based on priors on color perception.

\section{Competition of Unrestricted Adversarial Attacks on ImageNet}

The goal of this competition is to facilitate reliable evaluation of adversarial robustness of the current defense models in computer vision. In this competition, the organizers will collect various defense models on public benchmarks (e.g., CIFAR-10, ImageNet) for robustness evaluations. The competitors need to develop strong attack algorithms to find the worst-case robustness of those models. We aim to motivate novel attack algorithms to evaluate the adversarial robustness more effectively as well as evaluating the robustness of the current defenses reliably. 

\subsection{Dataset}
The competition uses the validation set of ImageNet-1K. We selected 15 images for each of the 1000 ImageNet classes according to the rule that the chosen image must be classified correctly by target models, resulting 15000 test images. Then the whole test set is splitted into three parts, used by three competition stages respectively.

\subsection {Tasks and Competition Rules}

The task for the contestants is designing an algorithm which modifies the test images offline to make the target models mis-classified. The whole competition is divided into three stages: first, we open 5000 images for players to debug the algorithm, the samples submitted by players will be predicted by the target models and scored by a series of pre-defined metrics; after the first stage, we open another 5000 images, and go into the second stage, in this stage the scoring metrics is consistent with stage I, but we replaced the target image recognition model. The second stage lasted only 7 days; After the end of the second stage, we will select the teams that enter the final stage through two principles: 1) According to the objective score of the second stage, the top 10 teams will enter the final stage; 2) According to the objective score of the second stage, the top 11-50 teams will be recommended anonymously by the experts of the organizing committee, and the top 10 teams will enter the final stage. In the final stage, players need to submit the model and running environment for generating adversarial examples. Finally, we will test on the remaining 5000 private images and employ labours to score the submitted unrestricted adversarial examples which can attack successfully. The scoring mechanism is considered from two dimensions: 1) the changed image needs to have the same semantics as the original image and guarantee the singleness of the semantics; 2) the image needs to have higher visual quality, and human can clearly identify the object. The final rank of the competition is decided by the manual scoring results.

\subsection{Evaluation Metrics}

\paragraph{Target models of stage I.} Three high-performance networks: Efficient-B5, ResneXt101-32$\times$8d and Inception-V4 are adopted as target models in stage I. They are from three branchs of network architecture design, and have better complementarity. For data preprocessing, in addition to the traditional \emph{center crop} and \emph{normalization}, we also add \emph{padding} and \emph{gaussian blur} to make the target models have a certain defensive ability.

\paragraph{Target models of stage II.} In stage II, we use three models with larger capability: Efficient-L2, ViT-Lagre and ResNeSt269. It will increase the difficulty of attack, and help for screening out better attack algorithms. Same with stage I, preprocessing-based defense is used at inference time.

\paragraph{Objective machine scoring metrics.} We use three rigorous evaluation metrics proposed in previous works as follows:
\begin{itemize}
\item Attack Success Rate (ASR). ASR is the proportion of examples attacking successfully in all test examples. We directly calculate the ASR of the adversarial examples submitted by players, which is in [0,1]:
\begin{equation}
S_{ASR}^{*}=\frac{\|\{\hat{x}|\mathcal{F}(\hat{x}) \neq y\}\|}{N},
\end{equation}
where $\mathcal{F}(\hat{x})$ is the output of the target model. $N$ is the total number of test images. 

\item Fréchet Inception Distance (FID). We limit the images submitted by competitors to satisfy the natural characteristics. Inspired by the evaluation metrics of GANs or other generative models, such as Inception Score (IS), FID and KID, we take FID as the metric of image naturalness:
\begin{equation}
S_{FID}^{*}=\sqrt{1-\frac{min(FID(X, \hat{X}),200)}{200}},
\end{equation}
where $X$ is the clean test image set and $\hat{X}$ stands for the set of adversarial examples submitted by competitors. We normalized the FID score into [0,1].

\item Learned Perceptual Image Patch Similarity (LPIPS). LPIPS is a measurement of perceptual distance, which is better than the traditional image quality metrics SSIM and PSNR. The main purpose of adding perceptual distance between $X$ and $\hat{X}$ is to prevent the contestants to artificially select some online public examples to submit. Such a practice can achieve high attack success rate, but is meaningless in this competition. The formulation of LPIPS is as follows:
\begin{equation}
    S_{LPIPS}=\sum _{l}\frac{1}{H_{l}W_{l}}\sum _{h,w}d(f^{l}_{hw},\hat{f}^{l}_{hw}),
\end{equation}
where $f^{l}_{hw}$ and $\hat{f}^{l}_{hw}$ is the $l$th layer feature output of VGG with the input of $X$ and $\hat{X}$. $d$ is the L2 distance metric. The value of $S_{LPIPS}$ also needs to be normalized by:
\begin{align}
    & Clip(S_{LPIPS}) = min(max(S_{LPIPS},0.2),0.7) \\
    & S_{LPIPS}^{*} = \sqrt{ 1-2*(Clip(S_{LPIPS})-0.2)}
\end{align}

\end{itemize}

Finally, the overall objective machine score is:
\begin{equation}
    S_{sub}=100 \times S_{ASR}^{*} \times S_{FID}^{*} \times S_{LPIPS}^{*}.
\end{equation}

\paragraph{Subjective human scoring metrics.} At the end of the competition, contestants need to submit their method to attack 5000 private images. The generated adversarial examples will be manually scored from two aspects: image semantic and quality. The final rank will be determined according to the results of the subjective scoring.

\begin{itemize}
\item Image semantics annotation. Image semantic annotation is to determine whether the image semantic changes before and after the attack. The attack is successful only if the image semantics is preserved and the model recognition is wrong.

The specific annotation steps are as follows: each annotator will be given the original image and corresponding adversarial examples. They will judge whether the semantics of the image after unrestricted attack has changed. Semantic annotation is more subjective. Each pair will be assigned to five people for annotation, and the score decided by the majority vote of five people will be used as the semantic score of the image. If the semantics of the submitted image changes, then $S_{s} = 0$, otherwise $S_{s}=1$.
\item Manual annotation of image quality. Image quality manual annotation is a quantitative annotation of the change of image quality before and after the attack, which is used to determine the quality of the generated unrestricted adversarial image.

The specific annotation steps are as follows: similar with image semantics annotation, the image quality will also be in the form of pair, and the degree of quality before and after the attack will be annotated. We quantify the image quality annotation, and divide it into five levels: 1) The image quality is the highest, and the original image is basically consistent with the image after the attack; 2) The image quality is good, and the object can be clearly identified; 3) The image quality looks so so; 4) The image has poor quality and can not be recognized at a glance; 5) The image quality is very poor, can barely identify the object. Each pair will be assigned 5 people to annotate, and finally the average marked score of 5 people will be used as the image quality score $S_{q}$.

\end{itemize}

The final total subjective evaluation score is as follows:
\begin{equation}
    S_{obj}=\sum_{\hat{x}}^{\mathcal{F}(\hat{x}) \neq y} S_{s}(\hat{x})\times \frac{S_{q}(\hat{x})}{5}.
\end{equation}
It should be noted that in the final 5000 images, we only count on the subjective scores of examples which can attack successfully.

\section{Competition Results}
By running and evaluating 3007 submissions from around one hundred teams, we select the top 10 teams to win the prize. The competition results of top 5 teams have been shown in Table~\ref{tab:analysis_other}. We also report the ASR and score in two extra stages before the final scoring. In this competition, ASR is not the determinant of getting a high final score. The contestants need to consider the quality of modified image, the generality and transferability of the attacks, etc. This setting leads the contestants to design a more practical unrestricted attack. Among the top 5 teams, GoodAdv has the lowest ASR and score in stage I and stage II, but it still gets the final score of second place, as the most modified images having good visual quality. The Top-1 team AdvRandom develops an attack algorithm with good transferability, which keeps high ASR and score in both stages. Finally AdvRandom gets a score of 2210, achieving the first place. For the description and implementation details of top 5 methods, we introduce them in Section~\ref{sec:method}.

\begin{table}[t]
\small
    \centering
     \tablestyle{5pt}{1.05}
\begin{tabular}{c|c|c|c|c|c}
\toprule

\multirow{2}{*}{Team}
& \multicolumn{2}{c|}{\textbf{Stage I}} 
& \multicolumn{2}{c|}{\textbf{Stage II}} & \multirow{2}{*}{Final} \\
 & ASR (\%) & Score  & ASR (\%) & Score & \\
\midrule

AdvRandom & 86.96 & 81.98 & 82.76 & 74.93 & 2210 \\ 
\midrule
GoodAdv & 46.08 & 42.40 & 76.04 & 57.32 & 2114 \\ 
\midrule
DemiguiseWoo & 92.78 & 84.69 & 76.6 & 63.31 & 1861 \\
\midrule
Advers & 95.14 & 87.80 & 80.0 & 66.27 & 1816 \\
\midrule
fangzheng00 & 78.26 & 71.17 & 80.0 & 56.37 & 1805 \\
\bottomrule

\end{tabular}
\vspace{1em}
    \caption{Competition results of Top-5 teams.}
    \label{tab:analysis_other}
\end{table}

\section{Top Scoring Submissions}
\label{sec:method}
%--------------------------------------------1st-------------------------------------------
%\hang{Top teams, Yinpeng, Hang}
\subsection{1st place: AdvRandom}
\textbf{Team members}: Fangcheng Liu (Peking University), Chao Zhang (Peking University), Hongyang Zhang (TTIC \& University of Waterloo).

\subsubsection {Method}
The final total subjective evaluation score is decided by two parts, i.e., attack success rate and \emph{image quality}. Therefore, it is important to reach high attack success rate while maintaining original semantic and good perceptual quality. However, a key challenge in computer vision is the lack of a precise mathematical metric of human perception \cite{laidlaw2021perceptual}. In this competition, we aim to find the smallest perturbation $\delta^*$ such that $x +\delta^*$ is misclassified by the target model $\mathcal{F}$ under specific distance metric $\|\cdot\|$, i.e., 
\begin{equation}
    \begin{aligned}
        \delta^* = & \argmin_{\delta} \; \|\delta\|, \\
        \mathrm{s.t.}\;\; & \mathcal{F}(x +\delta) \neq y, \\
        & x +\delta \in [0, 1]^d.
    \end{aligned}
    \label{eq: argmin}
\end{equation}
However, direct optimization of problem \eqref{eq: argmin} is intractable, in part due to the lack of information about the target model $\mathcal{F}$. We approximately solve this problem by discretizing the continuous space of perturbation size into a discrete set and choosing the minimum perturbation size so that the attack is able to fool the target model $\mathcal{F}$. However, the challenge is that it is typically difficult to decide whether a given perturbation radius can also fool the target model~\cite{cheng2019improving}. This problem is also known as model selection problem, and a classic approach to tackle this problem is to use a validation model to help decide the proper perturbation radius. More specifically, we split all available models into training model set and validation model set. Adversarial examples are crafted on training model and we will stop increasing perturbation budget for further attack if the probability of the true class on the validation model is smaller than a certain threshold $\eta$. With the validation model, we are able to select the minimum perturbation size to fool the target model $\mathcal{F}$ from the discrete set of perturbation radius.

In this competition, we consider $\ell_{\infty}$ distance metric. Given a perturbation radius $\varepsilon$ from the discrete set, We formulate our transfer-based attack by combining Translation-invariant method~\cite{dong2019evading}, Diverse input method~\cite{xie2019improving}, and Momentum Iterative method~\cite{dong2018boosting} (TDMI), which can be formulated as
\begin{equation} \label{TDMI}
\begin{aligned}
    m_{t+1} & = \mu \cdot m_t + \frac{\mathbf{W} * \nabla_{\boldsymbol{x}_t}L\left(f\left(T(\boldsymbol{x}_t, p);\boldsymbol{\theta}\right), y\right)}{\|\mathbf{W} * \nabla_{\boldsymbol{x}_t}L\left(f\left(T(\boldsymbol{x}_t, p);\boldsymbol{\theta}\right), y\right)\|_1}, \\
    \boldsymbol{x}_{t+1} & = \Pi_{\mathbb{B}(\boldsymbol{x}, \varepsilon)}\left(\boldsymbol{x}_{t} + \alpha \cdot \text{sign}(m_{t+1})\right),
\end{aligned}
\end{equation}
where $m_0 = 0, \boldsymbol{x}_{0} = \boldsymbol{x}$, $\mathbf{W}$ is a pre-defined kernel with a convolution operation $*$, $\alpha$ is the step size, and $\mu$ is the decay factor for the momentum term. $T(\boldsymbol{x}_t, p)$ represents the diverse input transformation on $\boldsymbol{x}_t$ with probability $p$. 

\subsubsection{Submission Details and Results}
In our final submission, we set $\mu = 0.8, p = 0.7, \eta = 0.01$ and consider $\varepsilon \in \{4, 6, 8, 12, 16, 32, 64\}$. The kernel size of $\mathbf{W}$ is set to $5 \times 5$. Our training and validation models are both an ensemble~\cite{liu2016delving} of eight high-performance networks. %\hang{The details is a little bit complex}
% For training model set, We use Vit-deit-base-distilled, dm-nfnet-f1, Efficient-B4-NS, Efficient-B5-NS, Resnet269, Resnet152-V2, ResneXt101-32×16d, Inception-V4; For validation model set, we use Resnet50-V2, Resnet101-V2, Inception-V3, Inception-Resnet-v2, Cspdarknet53, Densenet201, Repvgg-b2g4, Dpn107.

The final total subjective evaluation score of our approach is 2210, surpassing the runner-up by 4.56\%. Besides, we can define the average image quality score\footnote{The upper bound of average image quality score is 5} by
\begin{equation}
    S_{quality}= \frac{S_{obj}}{1000 \times S_{ASR}^{*} },
\end{equation}
where $S_{obj}$ is the final total subjective evaluation score and $S_{ASR}^{*}$ is the attack success rate. The average image quality score of our approach is 2.85, surpassing the runner-up 2.3 by 23.91\%.
%--------------------------------------------1st-------------------------------------------

%--------------------------------------------2st-------------------------------------------
\subsection{2nd place: team GoodAdv}
\textbf{Team members}:  Xiao Yang, Yichi Zhang, Shilong Liu, Chang Liu, Wenzhao Xiang 

\subsubsection {Method}
%The traditional adversarial attacks calculate the gradient to each one of the image pixel~\cite{madry2019deep,goodfellow2015explaining}, leading to a noise-like perturbations on the input image. However, the noise-like perturbations can be easily caught by human eyes, which means the perturbation amplitude should be limited to obtain higher image quality. Inspired by the fact that human eyes are not sensitive to high frequency signal changes, we intend to deal with this problem from frequency domain.
Although the adversarial perturbations generated by the existing methods (e.g., PGD~\cite{kurakin2016adversarial} and MIM~\cite{dong2018boosting}) have a small intensity change (e.g., $16$ for each pixel in $[0,255]$), they may still sacrifice the visual quality for human perception, and a similar conclusion is also elaborately described in~\cite{sen2019should} that $\ell_{p}$-norm adversarial perturbations can not fit human perception well. To make the crafted image indistinguishable from the corresponding original one, we require that the crafted image should look natural beyond the constraint of the $\ell_p$ norm bound. Inspired by the intuitive fact that human eyes are not sensitive to high frequency information, we aim to tackle this problem 
in the frequency domain.

Specifically, we separately manipulate different frequency bands of the input image to make more adversarial perturbations appear in high frequency bands, indicating explicitly encouraging the naturalness of the generated image. Therefore, given the input image $\bm{x}$ and the corresponding label $y$, the optimization problem can be formulated as

\begin{equation}
\label{optimization}
    W^*=\mathop{\max}\limits_{W}\mathcal{J}\Big(f\big(\mathcal{F}^{-1}(W\odot \mathcal{F}(\bm{x}))\big),y\Big),
\end{equation}
where $f$ represents the classifier, $\mathcal{J}$ is loss function of the classifier (e.g., cross-entropy loss and CW loss~\cite{carlini2017evaluating}). And $\mathcal{F}$ and $\mathcal{F}^{-1}$ represent the corresponding forward and inverse frequency transformation, $W$ indicates the attention map of the coefficients in frequency domain. The optimized adversarial samples can be obtained from the optimal attention map $W^*$ in Eq.~\eqref{optimization}. Therefore, the generation process can be expressed as
\begin{equation}
\label{sample}
    \bm{x}^{adv}=\mathcal{F}^{-1}(W^* \odot \mathcal{F}(\bm{x})).
\end{equation}

In order to make the crafted image both perceptually satisfactory and effective against different classifiers, we devise the dynamic learning rate of the frequency coefficients among different frequency bands. Specifically, the coefficients in higher frequency bands will be manipulated with a higher learning rate, while a lower learning rate will be adopted in lower frequency bands.

As described in the optimization formulation Eq.~\eqref{optimization}, the optimization process tends to explore the attention map. The design of the attention map is inspired by the filters in signal processing. Since any signal can be expressed as a combination of signals with different frequencies, the attention map can be used to rearrange the weights of these signals. We adopt Fourier transformation as the frequency domain transformation, and the forward transformation can be expressed as
\begin{equation}
\label{ft}
    F(u,v)=\sum_{i=0}^{M-1} \sum_{j=0}^{N-1}I_{\bm{x}}(i,j)e^{-j2\pi(\frac{ui}{M}+\frac{vj}{N})},
\end{equation}
where $M$ and $N$ represent the sampling numbers in two directions, $I_{\bm{x}}(i,j)$ is the original image $\bm{x}$ in spatial domain, and $F(u,v)$ calculates corresponding values in frequency domain. Rather than directly reconstructing the input image by inverse Fourier transformation, we introduce an attention map into the coefficients in frequency domain as
\begin{equation}
\small
\label{ift}
\begin{split}
    I_{\bm{x}^{adv}}(i,j)=\mathrm{Re}(\frac{1}{MN}\sum_{u=0}^{M-1} \sum_{v=0}^{N-1}W(u,v)
    \cdot F(u,v)e^{j2\pi(\frac{ui}{M}+\frac{vj}{N})}),
\end{split}
\end{equation}
where $W(u,v)$ is the optimized attention map, and $\mathrm{Re}(\cdot)$ means taking the real part.

\subsubsection{Submission Details and Results}
\begin{figure}[tp]
  \centering
  \includegraphics[width=1.\columnwidth]{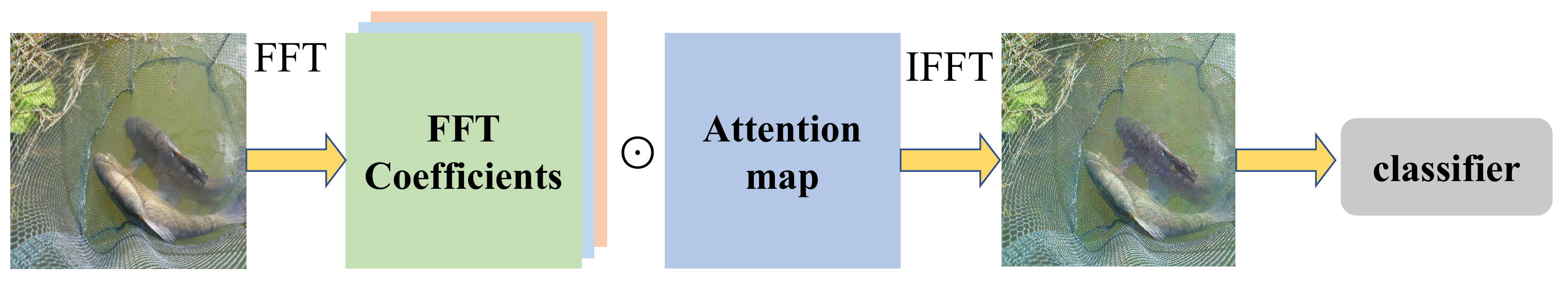} 
  \caption{The pipline of the frequency-based adversarial attack.}
  \label{fig:1}
  \vspace{-2ex}
\end{figure}
The whole pipeline is illustrated in Figure~\ref{fig:1}. First, the input image is transformed into frequency domain by Fourier transform. Then the coefficients matrix in the frequency domain is multiplied by an attention map,  and the manipulated coefficients are transformed to spatial domain by inverse Fourier transform. Finally, the manipulated image is fed into the classifier to calculate the gradients w.r.t the attention map. We present some details as follows

\begin{itemize}
\item The attention map is built as $1\times M \times N$ matrix, meaning that the attention map is shared across channels. The reason for sharing maps is that color manipulations will lead to perceptible evidence. After sharing the attention map across channels, the color difference can be limited to an acceptable range.
\item The gradient map of the loss is normalized with a minus Gaussian kernel, indicating that lower frequency coefficients have smaller learning rates, the higher frequency coefficients have bigger learning rates. The weight kernel can be optional. We observe that the Gaussian kernel shows better visual quality than the linear kernel in our experiments. 
\item We select CW loss as the loss function. 
%The advantage of using CW loss is that we can continue the attack process even if the attack has already succeeded.
\end{itemize}

The final score of our method is 10567.8.
%--------------------------------------------2st-------------------------------------------

%--------------------------------------------3st-------------------------------------------

\subsection{3rd place: team DemiguiseWoo}
\textbf{Team members}: Yajie Wang, Huipeng Zhou, Haoran Lyu, Yidan Xu, Zixuan Xu, Taoyu Zhu, Wenjun Li, Xianfeng Gao

In this section, we introduce the attack method we use, which won third place in the Competition of Unrestricted Adversarial Attacks on ImageNet. We describe our approach in~\ref{DemiguiseWooMethod}, and detail our submissions in~\ref{DemiguiseWooSubmission}.

\subsubsection {Method}
\label{DemiguiseWooMethod}
Our approach aims to generate unrestricted perturbations while ensuring that adversarial examples are friendly to the human visual system (HVS). Therefore, we design our objective function in two parts: the first part is the adversarial loss to guarantee the attack strength of adversarial examples. We use the cross-entropy loss between the predicted label of local models and the clean label. The second part is the Perceptual loss to guarantee the visual quality of adversarial examples. We use Learned Perceptual Image Patch Similarity (LPIPS) and pixel-level Mean Square Error (MSE). LPIPS is a metric that utilizes deep features from trained Convolutional Neural Networks to measure image similarity~\cite{zhang2018unreasonable}, and we compute LPIPS between the original image and the adversarial example. MSE considers the image realism loss problem and aims to reduce the data distribution distance between the adversarial example and the original image~\cite{shen2017ape}. With MSE, adversarial examples are generated as consistent as possible with the distribution of clean samples. Our loss function can be formulated as follows:

\begin{equation}
\mathcal{L}=\mathcal{L}_{ce}\left(\boldsymbol{x}^{adv}, y_{true}\right)+\lambda \cdot \mathcal{D}\left(\boldsymbol{x}^{adv}, \boldsymbol{x}\right) + \phi \cdot \mathcal{L}_{mse}\left(\boldsymbol{x}^{adv}, \boldsymbol{x}\right).
\end{equation}

Where $\mathcal{L}_{ce}$ is the cross-entropy loss between the label of $\boldsymbol{x}^{adv}$ and the ground-truth label $y_{true}$, $\mathcal{D}$ is the Perceptual Similarity distance, $\mathcal{L}_{mse}$ is the pixel-level Mean Square Error loss.

Our attack combines two practical improvements built upon the basic iterative method (BIM)~\cite{kurakin2016adversarial}: TI-FGSM~\cite{dong2019evading} and MI-FGSM~\cite{dong2018boosting}. First, we translate images to attenuate overfitting of adversarial examples to the recognition region of local models, facilitating adversarial examples transfer to the black-box model. Next, we add a momentum term to stabilize the update direction of gradients during the iterative process, avoiding falling into local optima and overfitting to local models. We also utilize the variance-reduced gradient to generate adversarial examples~\cite{wu2018understanding}, which further improves the transferability of adversarial examples. Finally, we ensemble multiple models locally to generate adversarial examples for attacking the unknown black-box model.

An effective way to improve the robustness and transferability of adversarial examples is image variation. We perform the random variation of input diversity~\cite{xie2019improving} and random excision of input samples to obtain more diverse inputs and enhance the adversarial examples' transferability. Also, we use the Sobel operator to obtain the edge contour information of the image. We partially eliminate it, which mitigates the trade-off between attack strength and perturbation visibility and reduces perturbation visibility.

In summary, our approach can be formulated as follows:
\begin{equation}
g_{t+1}=\mu \cdot g_{t}+\frac{\nabla_{x} \mathcal{L}\left(x_{t}^{adv}, y_{true}\right)}{\left\|\nabla_{x} \mathcal{L}\left(x_{t}^{adv}, y_{true}\right)\right\|_{1}},
\end{equation}

\begin{equation}
x_{t+1}^{adv}=\operatorname{clip}\left\{x_{t}^{adv}+\alpha \cdot \operatorname{sign}\left[\mathcal{T} *\left(g_{t+1}\right)\right]\right\}.
\end{equation}

Where $t$ is the number of iterations, $\mu$ is the decay factor, $\mathcal{L}$ is the loss function, $y_{true}$ is the original label of the adversarial example $x^{adv}$, $\mathcal{T}$ is the translation operator. In each iteration, we clip the perturbations to avoid them being conspicuous.

\subsubsection{Submission Details and Results}
\label{DemiguiseWooSubmission}
We fuse logit activations of multiple models as the local ensemble model, assigning equal weights to each model. Specifically, we ensemble the following models: DenseNet-201~\cite{huang2017densely}, RepVGG~\cite{ding2021repvgg}, GhostNet~\cite{han2020ghostnet} and Vision Transformer (ViT)~\cite{yuan2021tokens}. We follow the original settings of the attacks~\cite{dong2019evading,dong2018boosting,wu2018understanding,xie2019improving} and fine-tune parameters based on experience. We attack images in the range [0-255] with the number of iterative rounds set to 50 and the step size set to 0.36 for the first 45 rounds and 0.072 for the last 5 rounds.

For calculating the distance loss of data distribution between adversarial examples and clean images, we also tried Fréchet Inception Distance (FID), Cosine similarity and Kullback-Leibler Divergence. They are effective but not ideal for the competition.
%--------------------------------------------3st-------------------------------------------

%--------------------------------------------4st-------------------------------------------
\subsection{4th place: team Advers}
\textbf{Team members}: Guoqiu Wang, Huanqian Yan, Ying Guo, Chaoning Zhang.

\subsubsection {Method}

There are many methods for unrestricted adversarial attacks, e.g., adding $L_{p}$ norm perturbation \cite{goodfellow2015explaining, madry2017towards}, generating adversarial images by GAN networks \cite{wang2019at-gan}, sticking adversarial patches \cite{liu2019perceptual, guo2021meaningful}. In the first stage, we carried out many different methods and decided to generate adversarial images by adding $L_{\infty}$ norm  perturbation, which has a higher machine score.

Attack success rate is the most important item for the score in three stages. We should improve the transferability of adversarial images to improve the ASR, because of the target models are unknown. Recently, many
methods have been proposed to improve the transferability, such as multi-model ensemble transferable attack,
input diversity (DI) \cite{xie2019improving}, translation-invariant attack (TI) \cite{dong2019evading}, and
momentum-based attack (MI) \cite{dong2018boosting}.

In addition to the attack success rate, $S_{FID}^{*}$ and $S_{LPIPS}^{*}$ are also important for the machine score in the first two stages. And the adversarial images need to have higher visual quality for human in the third stage.
% Translation-invariant attack and momentum-based attack can improve the tranferability effectively, but the adversarial perturbations generated by them are more perceptible for human. 
Translation-invariant iterative attack improves attack ability by smoothing the gradient of the image
with a pre-defined kernel $W$. With
the size of $W$ increasing, the adversarial perturbations will be more perceptible. We set an appropriate size of $W$ in our solution by considering transferability and visual quality comprehensively. Adversarial images generated by momentum-based attack have a higher transferability, but the adversarial perturbations are more perceptible for human. We designed a new method named Gradient Refining (for more details, please read our paper \cite{wang2021improving}), which can balance the transferability and visual quality of adversarial images, and put away the momentum-based method.

Input diversity method brings random gradients and
translation-invariant method magnifies the randomness.
Though the random gradients reduce the over-fitting and
make the adversarial examples more generalized and transferable, there exists random useless information of gradients,
which inhibits the generalization and transferability of adversarial examples to some extent. Our experiments
prove this hypothesis.

To reduce the effects of random useless gradients introduced by input diversity, we design a method named Gradient Refining, which can enhance the transferability of
adversarial examples efficiently. In the process of attacking an image, our method randomly transforms the image
with several times during each iteration and calculates their
respective gradients, then averages them to get a refined
gradient, which preserves the gradients helpful for the transferability and counteracts the useless gradients for the attack.

\begin{figure}[t]
\centering\includegraphics[width=0.475 \textwidth]{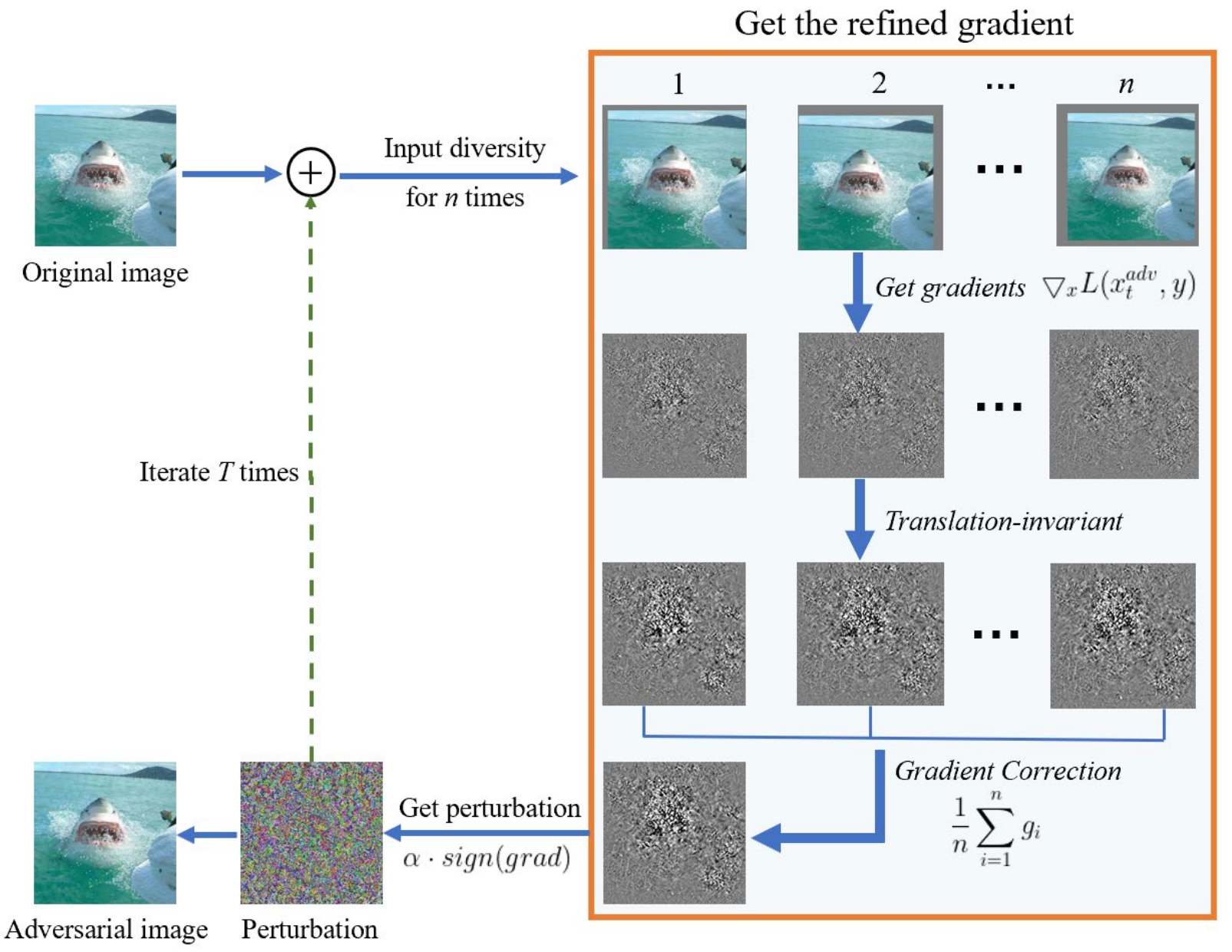}
\caption{\small The framework of R-DTI-FGSM algorithm. For more details, please read our paper \cite{wang2021improving}}
\label{fig:rng-dt}
\end{figure}

In our solution, we used R-DTI-FGSM algorithm (Gradient Refining combined with DI-TI), which can be formalized as:
\begin{equation}\label{eq:avg}
g_{r} = \frac{1}{n}\sum_{i=1}^{n} (W \ast \bigtriangledown_{x}L(T(x_{t}^{adv}, p), y))
\end{equation}

\begin{equation}\label{eq:rng}
x_{t+1}^{adv} = x_{t}^{adv} + \alpha \cdot sign(g_{r})
\end{equation}
where $\bigtriangledown _{x}L(T(x_{t}^{adv}, p), y)$ denotes the gradient of the loss function $L(\cdot, \cdot)$ w.r.t. the transformed image of adversarial example $x_{t}^{adv}$ generated in step $t$. $T(\cdot, p)$ denotes the stochastic transformation function with probability $p$. $y$ is the ground-truth label, $W$ denotes the pre-defined kernel and $n$ is the correction times of the gradient. $g_{r}$  is the refined gradient. $sign(\cdot)$ is the sign function and $\alpha$ denotes the step size. The framework of our solution is in Figure \ref{fig:rng-dt}.

\subsubsection{Submission Details and Results}
We set the  maximum perturbation of
each pixel to be 32/255, the step size $\alpha$ is 1/255 and the total iteration number $T$ is equal to $40$. For the stochastic transformation function $T(x,p)$, the probability $p$ is set to be $0.7$. The size of TI-kernel $W$ is $5 \times 5$. The correction times $n$ is set to be $9$. Ensemble models include ResNet50, DenseNet161, Inception-v4, Ens-adv-inception-resnet-v2, Efficientnet-b5-adv et al.

In the third stage, we won the fourth place in the final total subjective evaluation score (9081.6) and won the second place in attack success rate (95.48\%). 
%--------------------------------------------4st-------------------------------------------

%--------------------------------------------5st-------------------------------------------
\subsection{5st place: team fangzheng00}
\textbf{Team members}:Zheng Fang, Yang Wang, Bingyang Fu, Yunfei Zheng, Yekui Wang, Haorong Luo and Zhen Yang.

\subsubsection {Method}
\paragraph{} Our method is based on FGSM\cite{goodfellow2015explaining} and referred to Momentum Iterative boosting(MI)\cite{dong2018boosting}, Diversity Input(DI)\cite{xie2019improving} and Translation Invariant(TI)\textit{et al.}~\cite{dong2019evading} . Besides the transformations in MI, we used the rotation transformation to further enhance the transferability. Our method can be written as:
\begin{equation}
	x_{adv}^{t+1}=\Pi_{\epsilon}{x_{adv}^{t} + \alpha \cdot sign(W * \triangledown_{x}l(f(H(x_{adv}^{t})),y)}
\end{equation}

where $x_{adv}^{t} $ is adversarial example generated in iteration t and $W$ is a kernel.

The function $H$ includes three transformations($e.g.$, random size, random padding and rotation). The probability of transformation function is set to be 1 which means the input image would been transformed in every iteration. The rotation's angle is $-\pi/6$ to $\pi/6$. Other settings are the same as the original method.

We choose eight models, including four normal models, $i.e.,$Inception-V3\cite{christian2016rethinking}, Inception-V4\cite{christian2017inception}, Inception-Resnet-V2\cite{christian2017inception} and ResNet-V2 152\cite{he2016identity} and four adverarially trained models, $i.e.,$ ens3-adv-Inception-V3,	ens4-adv-Inception-V3, ens-adv-Inception-Resnet-V2 and adv-Inception-V3. The target model is an ensemble of above models by equal weight. The maximum perturbation is limited in $\epsilon$ = 16.
\subsubsection{Submission Details and Results}

\paragraph{}We find that increasing the number of iterations can get the adversarial examples with better transferability. Limited by time and hardware, the iteration is set to 100. In the final, we score was 9026.2 and attack success rate was 0.7838.
%--------------------------------------------5st-------------------------------------------

\section {Conclusion}
We introduce the Unrestricted Adversarial Attacks on ImageNet Competition. Though the elaborate design of rules and evaluation metrics, we obtain five practical and strong unrestricted adversarial attack algorithms, which has been described in Section~\ref{sec:method}. We wish this competition could give more inspiration to the community about practical unrestricted adversarial attacks. Furthermore, by collecting a set of attack algorithms, we will organize them to form a benchmark for evaluating the robustness of deep models against unrestricted adversarial attacks in the future.

{\small
\bibliographystyle{ieee_fullname}
\bibliography{egbib}
}

\end{document}